\documentclass[runningheads]{llncs}

\usepackage[T1]{fontenc}

\usepackage{cite}
\usepackage{amsmath,amssymb,amsfonts}
\usepackage{algorithmic}
\usepackage{graphicx}
\usepackage{textcomp}
\usepackage{xcolor}
\usepackage{hyperref}
\usepackage{subcaption}
\usepackage{xspace}
\usepackage{listings}
\usepackage{booktabs}
\def\BibTeX{{\rm B\kern-.05em{\sc i\kern-.025em b}\kern-.08em
    T\kern-.1667em\lower.7ex\hbox{E}\kern-.125emX}}

\newif\ifshowcomments
\showcommentstrue
\ifshowcomments
\newcommand{\mynote}[2]{\fbox{\bfseries\sffamily\scriptsize{#1}}
{\small$\blacktriangleright$\textsf{\emph{#2}}$\blacktriangleleft$}}
\else
\newcommand{\mynote}[2]{}
\fi

\newcommand{\ada}{AdaBoost.F\xspace}
\newcommand{\sys}{MAFL\xspace}
\newcommand{\intel}{Intel\textsuperscript{\textregistered}\xspace}

\newcommand\YAMLcolonstyle{\color{black}\mdseries\scriptsize}
\newcommand\YAMLkeystyle{\color{magenta}\bfseries\scriptsize}
\newcommand\YAMLvaluestyle{\color{black}\mdseries\scriptsize}
\makeatletter
\newcommand\language@yaml{yaml}
\expandafter\expandafter\expandafter\lstdefinelanguage
\expandafter{\language@yaml}
{
  keywordstyle=\color{magenta}\bfseries,
  basicstyle=\YAMLkeystyle,                                 
  sensitive=false,
  comment=[l]{\#},
  morecomment=[s]{/*}{*/},
  commentstyle=\color{purple}\ttfamily,
  stringstyle=\YAMLvaluestyle\ttfamily,
  moredelim=[l][\color{orange}]{\&},
  moredelim=[l][\color{magenta}]{*},
  moredelim=**[il][\YAMLcolonstyle{:}\YAMLvaluestyle]{:},   
  morestring=[b]',
  morestring=[b]",
  literate =    {---}{{\ProcessThreeDashes}}3
                {>}{{\textcolor{red}\textgreater}}1     
                {|}{{\textcolor{red}\textbar}}1
                {...}{{\mdseries\ \textcolor{black}{...}}}1
                {\ -\ }{{\mdseries\ \textcolor{black}{-}\ }}3,
}
\lst@AddToHook{EveryLine}{\ifx\lst@language\language@yaml\YAMLkeystyle\fi}
\makeatother
\newcommand\ProcessThreeDashes{\llap{\color{cyan}\mdseries-{-}-}}

\begin{document}

\title{Model-Agnostic Federated Learning}

\author{Gianluca Mittone\inst{1}\orcidID{0000-0002-1887-6911} \and
Walter Riviera\inst{2}\orcidID{0000-0001-5292-7594} \and
Iacopo Colonnelli\inst{1}\orcidID{0000-0001-9290-2017} \and
Robert Birke\inst{1}\orcidID{0000-0003-1144-3707} \and
Marco Aldinucci\inst{1}\orcidID{0000-0001-8788-0829}}

\authorrunning{G. Mittone et al.}

\institute{University of Turin, Turin, Italy \\
\email{\{gianluca.mittone, iacopo.colonnelli, robert.birke, marco.aldinucci\}@unito.it}
\and
University of Verona, Verona, Italy \\
\email{walter.riviera@univr.it}}


\begin{center}
The following paper is the accepted version of Springer copyrighted material
\\[12pt]
\textit{Mittone, G., Riviera, W., Colonnelli, I., Birke, R., Aldinucci, M. (2023). Model-Agnostic Federated Learning. In: Cano, J., Dikaiakos, M.D., Papadopoulos, G.A., Pericàs, M., Sakellariou, R. (eds) Euro-Par 2023: Parallel Processing. Euro-Par 2023. Lecture Notes in Computer Science, vol 14100. Springer, Cham}
\\[12pt]
presented at the EuroPar'23 conference in Limassol, Cyprus.
\\[12pt]
DOI: \href{https://doi.org/10.1007/978-3-031-39698-4_26}{10.1007/978-3-031-39698-4\_26}
\end{center}

\maketitle
\setcounter{footnote}{0}

\begin{abstract}
Since its debut in 2016, Federated Learning (FL) has been tied to the inner workings of Deep Neural Networks (DNNs); this allowed its development as DNNs proliferated but neglected those scenarios in which using DNNs is not possible or advantageous. 
The fact that most current FL frameworks only support DNNs reinforces this problem.
To address the lack of non-DNN-based FL solutions, we propose \sys (Model-Agnostic Federated Learning).
\sys merges a model-agnostic FL algorithm, \ada, with an open industry-grade FL framework: \intel OpenFL. 
\sys is the first FL system not tied to any machine learning model, allowing exploration of FL beyond DNNs.
We test \sys from multiple points of view, assessing its correctness, flexibility, and scaling properties up to 64 nodes of an HPC cluster.
We also show how we optimised OpenFL achieving a 5.5x speedup over a standard FL scenario. 
\sys is compatible with x86-64, ARM-v8, Power and RISC-V. 
\keywords{Machine Learning \and Federated Learning \and Federated {{AdaBoost}} \and Software Engineering}
\end{abstract}



\section{Introduction}

Federated Learning (FL) is a Machine Learning (ML) technique that has gained tremendous popularity in the last years~\cite{kairouz2021advances}: a shared ML model is trained without ever exchanging the data owned by each party or requiring it to be gathered in one common computational infrastructure.
The popularity of FL caused the development of a plethora of FL frameworks, e.g., Flower~\cite{beutel2020flower}, FedML~\cite{chaoyanghe2020fedml}, and HPE Swarm Learning~\cite{warnat2021swarm} to cite a few. 
These frameworks only support one ML model type: Deep Neural Networks (DNNs). 
While DNNs have shown unprecedented results across a wide range of applications, from image recognition~\cite{cnn} to natural language processing~\cite{SutskeverVL14}, from drug discovery~\cite{zhavoronkov2019deep} to fraud detection~\cite{KleanthousC20}, they are not the best model for every use case.
DNNs require massive amounts of data, which collecting and eventually labelling is often prohibitive; furthermore, DNNs are not well-suited for all types of data. 
For example, traditional ML models can offer a better performance-to-complexity ratio on tabular data than DNNs~\cite{o2019deep}.
DNNs also behave as black-box, making them undesirable when the model's output has to be explained~\cite{holzinger2019causability}.
Lastly, DNNs require high computational resources, and modern security-preserving approaches, e.g.~\cite{MeeseCALSN22,SotthiwatZLZ21}, only exacerbate this issues~\cite{lyu2022privacy}. 

We propose the open-source \textbf{\sys}\footnote{\url{https://github.com/alpha-unito/Model-Agnostic-FL}} (\emph{Model-Agnostic Federated Learning}) framework to alleviate these problems. 
MAFL leverages \emph{Ensemble Learning} to support and aggregate ML models independently from their type.
Ensemble Learning exploits the combination of multiple \textit{weak learners} to obtain a single \textit{strong learner}. 
A weak learner is a learning algorithm that only guarantees performance better than a random guessing model; in contrast, a strong learner provides a very high learning performance (at least on the training set).
Since weak learners are not bound to be a specific ML model, Ensemble Learning techniques can be considered \emph{model-agnostic}.
We adopt the \ada algorithm~\cite{polato2022boosting}, which leverages the AdaBoost algorithm~\cite{freund1997decision} and adapts it to the FL setting, and we marry it with an open-source industry-grade FL platform, i.e., \intel OpenFL~\cite{openfl}.
To our knowledge, MAFL is the first and only model-agnostic FL framework available to researchers and industry at publication.

The rest of the paper introduces the basic concepts behind \sys. 
We provide implementation details underlying its development, highlight the challenges we overcame, and empirically assess our approach from the computational performances and learning metrics points of view. 
To summarise, the contributions of this paper are the following:
\begin{itemize}
    \item we introduce \sys, the first FL software able to work with any supervised ML model, from heavy DNNs to lightweight trees;
    \item we describe the architectural challenges posed by a model-agnostic FL framework in detail;
    \item we describe how \intel OpenFL can be improved to boost computational performances;
    \item we provide an extensive empirical evaluation of \sys to showcase its correctness, flexibility, and performance. 
\end{itemize}

\section{Related Works}
\label{ssec:frameworks}
\emph{FL}~\cite{mcmahan2017communication} usually refers to a centralised structure in which two types of entities, a single \emph{aggregator} and multiple \emph{collaborators}, work together to solve a common ML problem. 
A FL framework orchestrate the federation by distributing initial models, collecting the model updates, merging them according to an aggregation strategy, and broadcasting back the updated model.
FL requires a \emph{higher-level software infrastructure} than traditional ML flows due to the necessity of exchanging model parameters quickly and securely. 
Model training is typically delegated to de-facto standard (deep) ML frameworks, e.g., PyTorch and TensorFlow.

Different \emph{FL frameworks} are emerging.
Riviera~\cite{Riviera2022} provides a compelling list of 36 open-source tools ranked by community adoption, popularity growth, and feature maturity, and Beltr{\'a}n~\cite{beltran2022decentralized} reviews 16 FL frameworks, identifying only six as mature. 
All of the surveyed frameworks support supervised training of DNNs, but only FATE~\cite{liu2021fate}, IBM-Federated~\cite{ibmfederated}, and NVIDIA FLARE~\cite{roth2022nvidia} offer support for a few different ML models, mainly implementing federated K-means or Extreme Gradient Boosting (XGBoost): this is due to the problem of defining a model-agnostic aggregation strategy. 
DNNs' client updates consist of tensors (mainly weights or gradients) that can be easily serialised and mathematically combined (e.g., averaged), as are also the updates provided by federated K-means and XGBoost.
This assumption does not hold in a model-agnostic scenario, where the serialisation infrastructure and the aggregation mechanism have to be powerful enough to accommodate different update types.
A truly model-agnostic aggregation strategy should be able to aggregate not only tensors, but also complex objects like entire ML model.
\ada is capable of doing that.
Section~\ref{sec:MAFA} delves deeper into the state-of-the-art of federated ensemble algorithms.


As a base for developing \sys, we chose a mature, open-source framework supporting only DNNs: \intel OpenFL~\cite{openfl}.
The reason for this choice is twofold: (i) its structure and community support; and (ii) the possibility of leveraging the existing ecosystem by maintaining the same use and feel.
Section~\ref{sec:arc} delves into the differences between plain OpenFL and its \sys extension, showing how much DNN-centric a representative modern FL framework can be.

\section{Model-agnostic Federated Algorithms}
\label{sec:MAFA}

\begin{figure*}[t]
     \centering
     \begin{subfigure}[b]{0.3\textwidth}
         \centering
         \includegraphics[width=\textwidth]{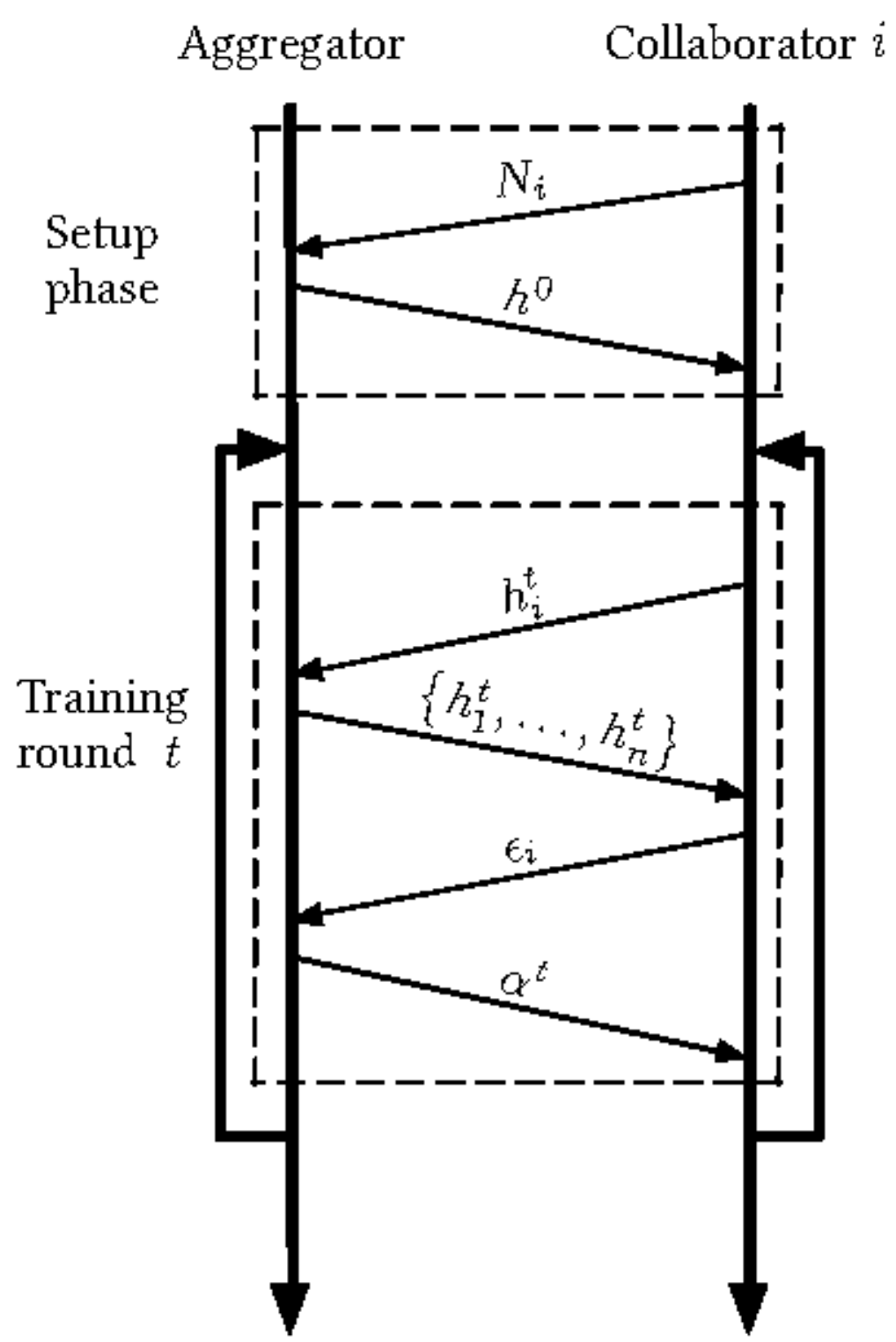}
         \caption{\textbf{DistBoost.F}}
         \label{fig:y equals x}
     \end{subfigure}
     \hfill
     \begin{subfigure}[b]{0.3\textwidth}
         \centering
         \includegraphics[width=\textwidth]{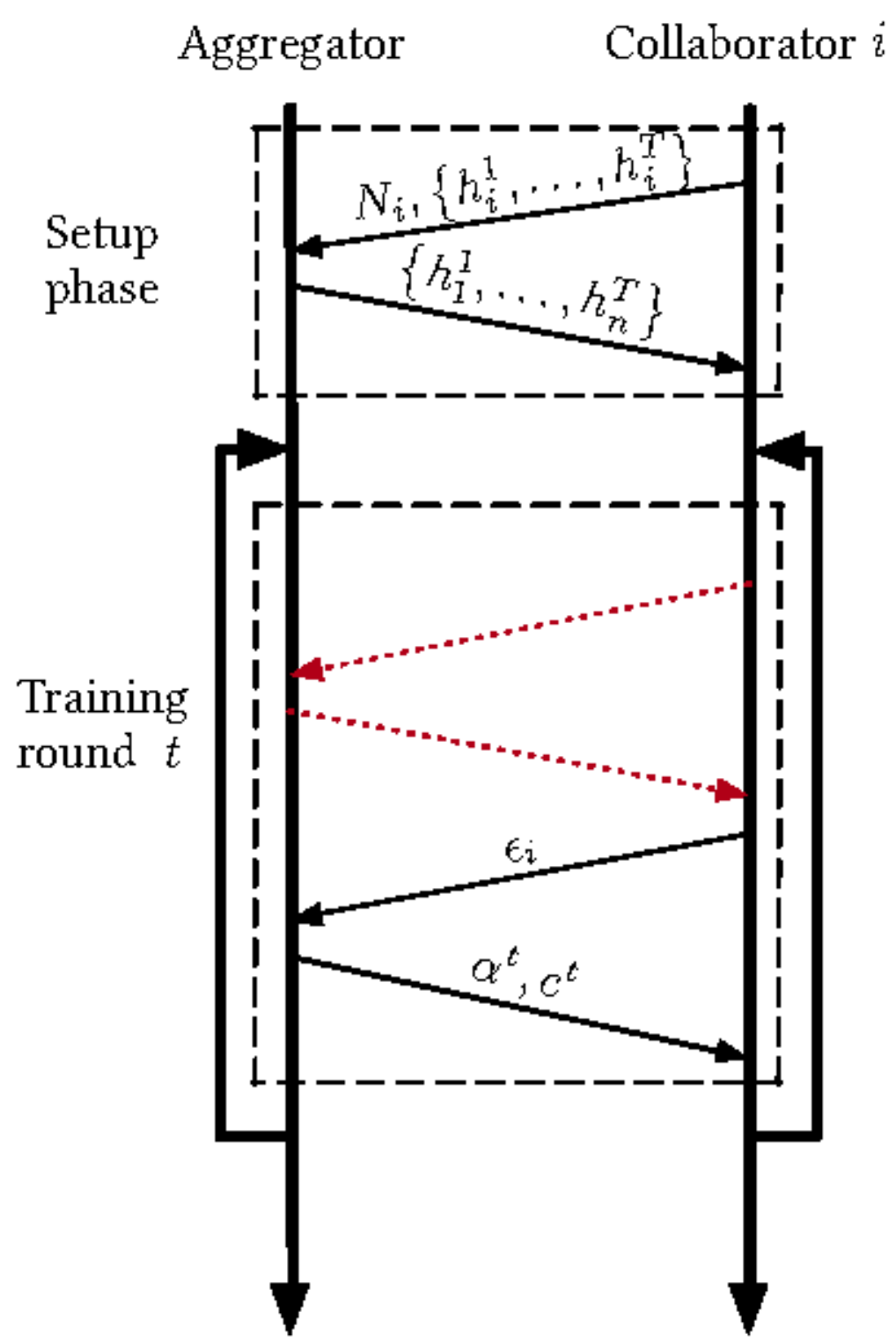}
         \caption{\textbf{PreWeak.F}}
         \label{fig:three sin x}
     \end{subfigure}
     \hfill
     \begin{subfigure}[b]{0.3\textwidth}
         \centering
         \includegraphics[width=\textwidth]{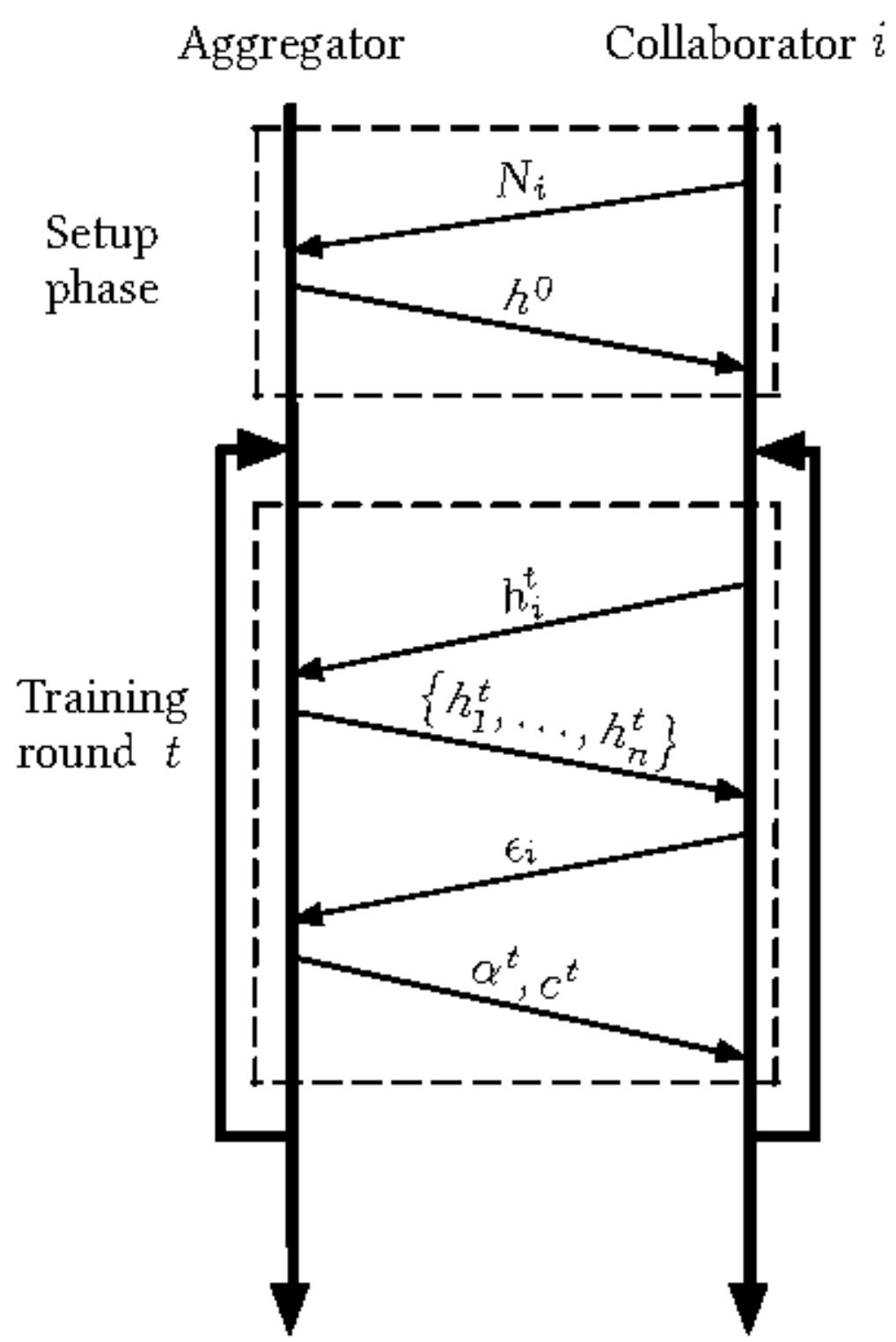}
         \caption{\textbf{AdaBoost.F}}
         \label{fig:five over x}
     \end{subfigure}
        \caption{\textbf{The three protocols implied by DistBoost.F, PreWeak.F, and AdaBoost.F}. $N$ is the dataset size, $T$ is the number of training rounds, $h$ the weak hypothesis, $\epsilon$ the classification error, $\alpha$ the AdaBoost coefficient. The subscript $i \in [1, n]$ indices the collaborators and the superscript $t$ the training rounds (with $0$ standing for an untrained weak hypothesis). $c \in [1, n]$ is the index of the best weak hypothesis in the hypothesis space. The red dotted line in PreWeak.F indicates the absence of communication.
        }
        \label{fig:federated_boosting}
\end{figure*}

None of the frameworks mentioned in Sec.~\ref{ssec:frameworks} supports model-agnostic FL algorithms, i.e., they cannot handle different ML models seamlessly.
The reason is twofold. 
On the one hand, modern FL frameworks still try to achieve sufficient technical maturity, rather than adding new functionalities. 
On the other hand, model-agnostic federated algorithms are still new and little investigated.

Recently, \cite{polato2022boosting} proposed three federated versions of AdaBoost: \emph{DistBoost.F}, \emph{PreWeak.F}, and \emph{AdaBoost.F}.
All three algorithms are model-agnostic due to their inherent roots in AdaBoost.
Following the terminology commonly used in ensemble learning literature, we call \emph{weak hypothesis} a model learned at each federated round and \emph{strong hypothesis} the final global model produced by the algorithms.
The general steps of an AdaBoost-based FL algorithm are the following:
\begin{enumerate}
    \item The aggregator receives the dataset size $N$ from each collaborator and sends them an initial version of the weak hypothesis.
    \item The aggregator receives the weak hypothesis $h_{i}$ from each collaborator and broadcasts the entire hypothesis space to every collaborator.
    \item The errors $\epsilon$ committed by the global weak hypothesis on the local data are calculated by each client and sent to the aggregator.
    \item The aggregator exploits the error information to select the best weak hypothesis $c$, adds it to the global strong hypothesis and sends the calculated AdaBoost coefficient $\alpha$ to the collaborators.
\end{enumerate}
Note that $N$ is needed to adequately weight the errors committed by the global weak hypothesis on the local data, thus allowing to compute $\alpha$ correctly.

Figure~\ref{fig:federated_boosting} depicts the protocol specialisations for the three algorithms described in~\cite{polato2022boosting}. 
They are similar once abstracted from their low-level details. 
While step 1 is inherently a setup step, steps 2-4 are repeated cyclically by DistBoost.F and AdaBoost.F. 
PreWeak.F instead fuses steps 1 and 2 at setup time, receiving from each collaborator $T$ instances of already trained weak hypotheses (one for each training round) and broadcasting $n\times T$ models to the federation. 
Then, each federated round $t$ loops only on steps 3 and 4 due to the different \emph{hypothesis space} the algorithms explore. 
While DistBoost.F and AdaBoost.F create a weak hypothesis during each federated round, PreWeak.F creates the whole hypothesis space during step 2 and then searches for the best solution in it.

All three algorithms produce the same strong hypothesis and AdaBoost model, 
but they differ in the selection of the best weak hypothesis at each round:
\begin{itemize}
    \item DistBoost.F uses a committee of weak hypotheses;
    \item PreWeak.F uses the weak hypotheses from a fully trained AdaBoost model;
    \item AdaBoost.F uses the best weak hypothesis trained in the current round.
\end{itemize}

The generic model-agnostic federated protocol is more complex than the standard FL one. 
It requires one more communication for each round and the exchange of complex objects across the network (the weak hypotheses), impacting performance. 
Note that each arrow going from collaborator $i$ to the aggregator in Fig.~\ref{fig:federated_boosting} implies a synchronisation barrier among all the collaborators in the federation. 
Increasing the number of global synchronisation points reduces concurrency and increases the sensitivity to stragglers.
It is worth noting that once an FL framework can handle the common protocol structure, implementing any of the three algorithms requires the same effort.
For this study, we implemented AdaBoost.F for two main reasons. 
First, its protocol covers the whole set of messages (like DistBoost.F), making it computationally more interesting to analyse than PreWeak.F. 
Besides, AdaBoost.F achieves the best learning results out of the three, also when data is heavily non-IID across the collaborators.

\section{\sys Architecture}
\label{sec:arc}

\begin{figure}[t]
  \centering
  \includegraphics[width=.8\linewidth]{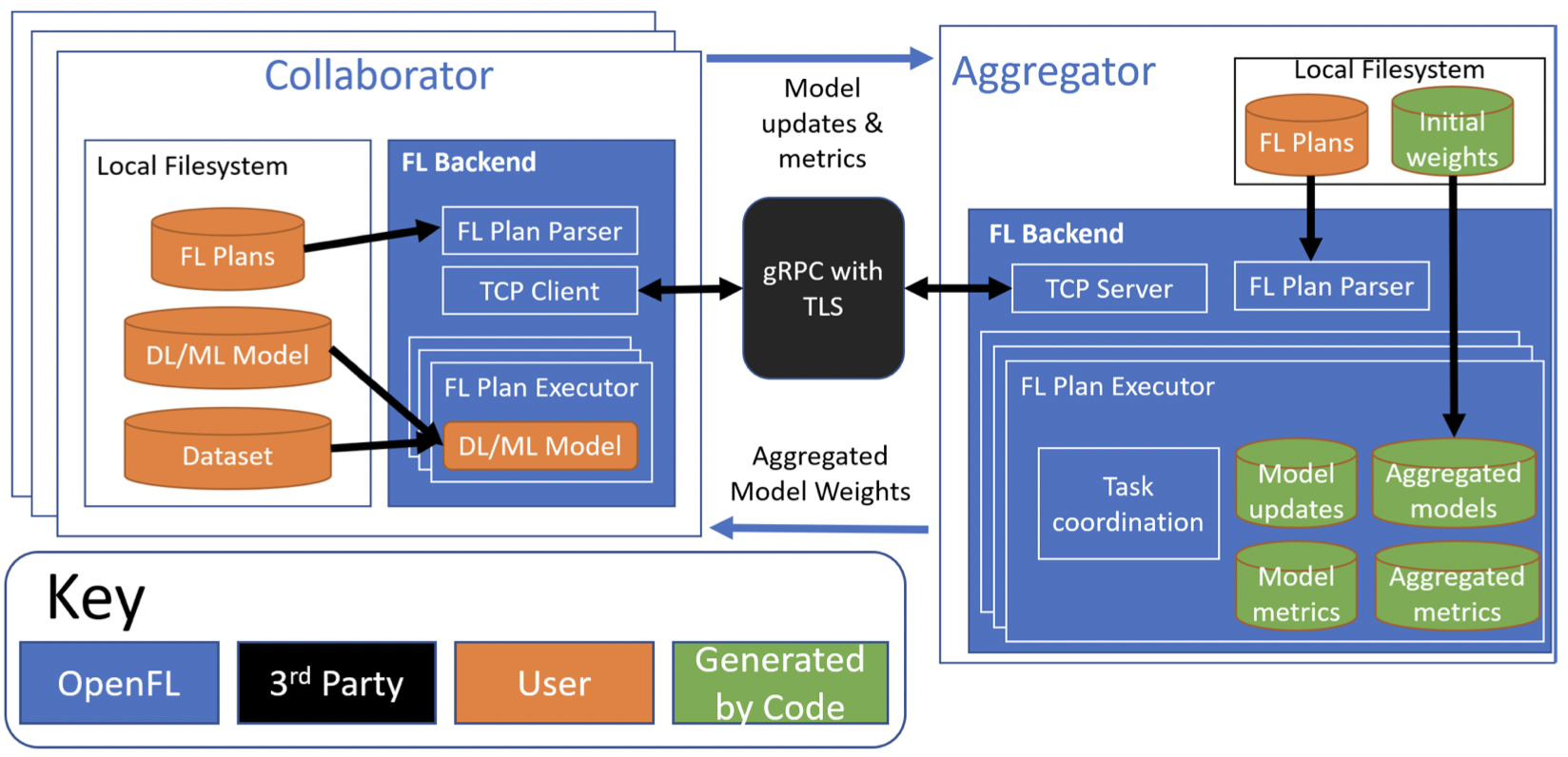}
  \caption{OpenFL architecture from~\cite{openfl}. The proposed extension targets only the inner components (coloured in blue).}
  \label{fig:openfl}
\end{figure}

Redesigning OpenFL comprises two main goals: allowing more flexible protocol management and making the whole infrastructure model agnostic.
During this process, we aimed to make the changes the least invasive and respect the original design principles whenever possible (see Fig.\ref{fig:openfl}).

\subsection{The Plan Generalization}
The \emph{Plan} guides the software components' run time.
It is a YAML file containing all the directives handling the FL learning task, such as which software components to use, where to save the produced models, how many rounds to train, which tasks compose a federated round, and so on. 
The original OpenFL Plan is rather primitive in its functions. 
It is not entirely customisable by the user, and many of its fields are overwritten at run time with the default values.
Due to its unused power, the parsing of the plan file has been extended and empowered, making it capable of handling new types of tasks, along with a higher range of arguments (and also making it evaluate \emph{every} parameter in the file).

The new model-agnostic workflow can be triggered by specifying the \texttt{nn: False} argument under the \texttt{Aggregator} and \texttt{Collaborator} keywords. 
The specific steps of the protocol can then be specified in the \texttt{tasks} section.
In the \intel OpenFL framework, there are only three possible tasks: 
\begin{itemize}
  \item \texttt{aggregated\_model\_validation}: test set validation of aggregated model;
  \item \texttt{train}: local training of the model;
  \item \texttt{locally\_tuned\_model\_validation}: test set validation of local model.
\end{itemize}
The three tasks are executed cyclically, with the Aggregator broadcasting the aggregated model before the first task and gathering the local models after the training step.
In \sys, the tasks vocabulary comprises three additional tasks:
\begin{itemize}
  \item \texttt{weak\_learners\_validate}: test set validation of the weak learners;
  \item \texttt{adaboost\_update}: update of the global parameters of AdaBoost.F on the Collaborators and the ensemble model on the Aggregator;
  \item \texttt{adaboost\_validate}: local test set validation of the aggregated AdaBoost.F model.
\end{itemize}
The \texttt{weak\_learners\_validate} task is similar to \texttt{aggregated\_model\-\_validation}. However, it returns additional information for AdaBoost.F, such as which samples are correctly predicted/mispredicted and the norm of the samples' weights.

The extended set of tasks allows users to use new FL algorithms, such as AdaBoost.F.
Additionally, if the \texttt{adaboost\_update} task is omitted, it is possible to obtain a simple \emph{Federated Bagging} behaviour. 
Switching behaviour requires small actions other than changing the Plan; however, both functionalities are documented with tutorials in the code repository. 

\subsection{Expanded Communication Protocol}
New messages have been implemented into the original \emph{communication protocol}, allowing the exchange of values other than ML models and performance metrics since \ada relies on exchanging locally calculated parameters.
Furthermore, \intel OpenFL only implements two synchronisation points in its original workflow: one at the end of the federation round and one when the Collaborator asks the Aggregator for the aggregated model.
These synchronisation points are hard-coded into the software and cannot be generalised for other uses.

For the AdaBoost.F workflow, a more general synchronisation point is needed: not two consecutive steps can be executed before each Collaborator has concluded the previous one.
Thus a new \texttt{synch} message has been added to the \emph{gRPC} protocol.
The working mechanism of this synchronisation point is straightforward: the collaborators ask for a \texttt{synch} at the end of each task, and if not all collaborators have finished the current task, it is put to sleep; otherwise, it is allowed to continue to the next task.
This solution, even if not the most efficient, respects the \intel OpenFL internal synchronisation mechanisms and thus does not require any different structure or new dependency.

\subsection{Core Classes Extension} 
The following core classes of the framework have been modified to allow the standard and model-agnostic workflows to coexist (see Fig.~\ref{fig:openfl} for an overview).

The \texttt{Collaborator} class can now offer different behaviours according to the ML model used in the computation. 
Suppose the Plan specifies that the training will not involve DNNs. 
In that case, the Collaborator will actively keep track of the parameters necessary to the AdaBoost.F algorithm, like the mispredicted examples, the weight associated with each data sample, and the weighted error committed by the models.
Additionally, the handling of the internal database used for storage will change behaviour, changing tags and names associated with the entries to make possible finer requests to it.

The \texttt{Aggregator} can now generate any ML models (instead of only DNNs weights), handle aggregation functions instantiated dynamically from the plan file, and handle the synchronisation needed at the end of each step.
New methods allow the Aggregator to query the internal database more finely, thus allowing it to read and write ML models with the same tags and name as the Collaborator.

\texttt{TensorDB}, the internal class used for storage, has been modified to accommodate the new behaviours described above.
This class implements a simple \emph{Pandas} data frame responsible for all model storage and retrieving done by the Aggregator and Collaborators. 
Furthermore, its \texttt{clean\_up} method has been revised, making it possible to maintain a fixed amount of data in memory. 
This fix has an important effect on the computational performance since the query time to this object is directly proportional to the amount of data it contains.

Finally, the more high-level and interactive classes, namely \texttt{Director} and \texttt{Envoy}, and the serialization library have been updated to work correctly with the new underlying code base.
These software components are supposed to be long-lived: they should constantly be running on the server and clients' hosts. 
When a new experiment starts, they will instantiate the necessary \texttt{Aggregator} and \texttt{Collaborators} objects with the parameters for the specified workflow.

This effort results in a model-agnostic FL framework that supports the standard DNNs-based FL workflow and the new AdaBoost.F algorithm.
Using the software in one mode or another does not require any additional programming effort from the user: a few simple configuration instructions are enough.
Additionally, the installation procedure has been updated to incorporate all new module dependencies of the software.
Finally, a complete set of tutorials has been added to the repository: this way, it should be easy for any developer to get started with this experimental software.

\section{Evaluation}

The complete set of tutorials replicating the experiments from~\cite{polato2022boosting} are used to assess \sys's correctness and efficiency.
We run them on a cloud and HPC infrastructure, both x86-64 based, and Monte Cimone, the first RISC-V based HPC system; however, \sys runs also on ARM-v8 and Power systems.

\subsection{Performance Optimizations}
\label{ssec:ablation}

\begin{figure}[t]
  \centering
  \includegraphics[width=.8\linewidth]{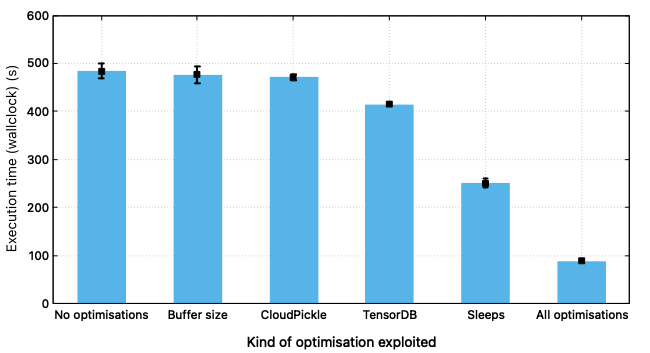}
  \caption{Ablation study of the proposed software optimisations; the 95\% CI has been obtained over five executions.}
  \label{fig:ablation}
\end{figure}

Using weak learners instead of DNNs drastically reduces the computational load. As an example, \cite{23:praise-fl:pdp} reports 18.5 vs 419.3 seconds to train a 10-leaves decision tree or a DNN model, respectively, on the PRAISE training set (with comparable prediction performance). Moreover, \ada requires one additional communication phase per round.
This exacerbates the impact of time spent in communication and synchronisation on the overall system performance. To reduce this impact, we propose and evaluate different optimisations to reduce this overhead. Applying all proposed optimisations, we achieve a 5.5x speedup on a representative FL task (see Fig.~\ref{fig:ablation}).
As a baseline workload,  we train a 10-leaves decision tree on the Adult dataset over 100 rounds using 9 nodes (1 aggregator plus 8 collaborators).
We use physical machines to obtain stable and reliable computing times, as execution times on bare-metal nodes are more deterministic than cloud infrastructures. 
Each HPC node is equipped with two 18-core \intel Xeon E5-2697 v4 @2.30 GHz and 126 GB of RAM.
A 100Gb/s \intel Omni-Path network interface (in IPoFabric mode) is used as interconnection network.
Reported times are average of five runs $\pm$ the 95\% CI.

We start by measuring the execution time given by the baseline: 484.13$\pm$15.80 seconds.
The first optimisation is to adapt the buffer sizes used by gRPC to accommodate larger models and avoid resizing operations. Increasing the buffer from 2MB to 32MB using decision trees reduced the execution time to 477.0$\pm$17.5 seconds, an improvement of $\sim1.5\%$. While this seems small, the larger the models, the bigger the impact of this optimisation.
The second optimisation is the choice of the serialisation framework: by using {\tt Cloudpickle}, we reduce the execution time to 471.4$\pm$6.1 seconds, an improvement of $\sim$2.6\%.
Next, we examine \texttt{TensorDB}, which grows linearly in the number of federated rounds, thus slowing down access time linearly.
We modified the \texttt{TensorDB} to store only the essential information of the last two federation rounds: this results in a stable memory occupation and access time.
With this change, the execution time drops to 414.8$\pm$0.9 seconds, an improvement of $\sim$14.4\% over the baseline.

Lastly, two \texttt{sleep} are present in the \sys code: one for the end-round synchronisation and another for the \texttt{synch} general synchronisation point, fixed respectively at 10 and 1 seconds.
Both have been lowered to 0.01 seconds since we assessed empirically that this is the lowest waiting time still improving the global execution time.
This choice has also been made possible due to the computational infrastructures exploited in this work; it may not be suitable for wide-scale implementations in which servers and clients are geographically distant or compute and energy-constrained.
With this sleep calibration, we obtained a global execution time of 250.8$\pm$9.6 seconds, a $\sim$48.2\% less than the baseline. Overall, with all the optimisations applied together, we can achieve a final mean execution time of 88.6$\pm$ seconds, i.e. a 5.46x speedup over the baseline.

\subsection{Correctness}
\label{ssec:correctness}

We replicate the experiments from~\cite{polato2022boosting} and compare the ML results.  
These experiments involve ten different datasets: \texttt{adult}, \texttt{forestcover}, \texttt{kr-vs-kp}, \texttt{splice}, \texttt{vehicle}, \texttt{segmentation}, \texttt{sat}, \texttt{pendigits}, \texttt{vowel}, and \texttt{letter}.
These are standard ML datasets targeting classification tasks, both binary (\texttt{adult}, \texttt{forestcover}, \texttt{kr-vs-kp}) and multi-class (all the others), with a varying number of features (from the 14 of \texttt{adult} up to the 61 of \texttt{splice}), and a different number of samples (from the 846 of \texttt{vehicle} up to the 495.141 of \texttt{forestcover}).
Each training set has been split in an IID way across all the Collaborators, while the testing has been done on the entire test set.
A simple Decision Tree from SciKit-Learn with ten leaves is used as a weak learner; instead, the AdaBoost class has been created manually.
We set the number of federated rounds to 300 and use 10 nodes: 1 aggregator plus 9 collaborators. We note that these optimizations can also benefit the original OpenFL.


\begin{table}[t]
    \centering
    \caption{Mean F1 scores $\pm$ standard deviation over 5 runs.}
    \label{tab:res}
    \begin{tabular}{lccr}
    \toprule
    \textbf{Dataset} & \multicolumn{1}{c}{\textbf{Classes}} & \multicolumn{1}{c}{\textbf{Reference}} & \multicolumn{1}{c}{\textbf{\sys}} \\
    \midrule
    Adult            & 2                                     & 85.58 ± 0.06                                  & 85.60 ± 0.05                                            \\
    ForestCover      & 2                                     & 83.67 ± 0.21                                  & 83.94 ± 0.14                                           \\
    Kr-vs-kp         & 2                                     & 99.38 ± 0.29                                  & 99.50 ± 0.21                                            \\
    \midrule
    Splice           & 3                                     & 95.61 ± 0.62                                  & 96.97 ± 0.65                                           \\
    Vehicle          & 4                                     & 72.94 ± 3.40                                  & 80.04 ± 3.30                                           \\
    Segmentation     & 7                                     & 86.07 ± 2.86                                  & 85.58 ± 0.06                                           \\
    Sat              & 8                                     & 83.52 ± 0.58                                  & 84.89 ± 0.57                                           \\
    Pendigits        & 10                                    & 93.21 ± 0.80                                  & 92.06 ± 0.44                                           \\
    Vowel            & 11                                    & 79.80 ± 1.47                                  & 79.34 ± 3.31                                           \\
    Letter           & 26                                    & 68.32 ± 1.63                                  & 71.13 ± 2.02                                           \\
    \bottomrule
    \end{tabular}
\end{table}

Table~\ref{tab:res} reports each dataset's reference and calculated F1 scores (mean value $\pm$ the standard deviation over five runs).
The values reported are fully compatible with the results reported in the original study, thus assessing the correctness of the implementation.
In particular, it can be observed that the standard deviation intervals are particularly high for the \texttt{vehicle}, \texttt{segmentation}, and \texttt{vowel}.
This fact can be due to the small size of the training set of these datasets, respectively 677, 209, and 792 samples, which, when split up across ten Collaborators, results in an even smaller quantity of data per client: this can thus determine the creation of low-performance weak learners.
Furthermore, also \texttt{letter} reported a high standard deviation: this could be due to the difference between the classification capabilities of the employed weak learner (a 10-leaves Decision Tree) compared to the high number of labels present in this dataset (26 classes), thus making it hard to obtain high-performance weak learners.

The mean F1 score curve for each dataset can be observed in Figure~\ref{fig:wandb}. 
As can be seen, after an initial dip in performance, almost each learning curve continues to grow monotonically to higher values.
This fact is expected since the AdaBoost.F is supposed to improve its classification performance with more weak learners.
It has to be observed that, at each federated round, a new weak learner will be added to the aggregated model: the AdaBoost.F grows linearly in size with the number of federated rounds.
This characteristic of the algorithm has many consequences, like the increasingly longer time needed for inference and for moving the aggregated model over the network.
From Figure~\ref{fig:wandb}, we can observe that, in the vast majority of cases, a few tens of federated rounds are more than enough to obtain a decent level of F1 scores; this is interesting since it is possible to obtain a small and efficient AdaBoost.F model in little training effort.
Instead, for the more complex datasets like \texttt{letter} and \texttt{vowel}, we can observe that it is possible to obtain better performance with longer training efforts.
This means that is possible to use AdaBoost.F to produce bigger and heavier models at need, according to the desired performance and inference complexity.


\subsection{Flexibility}
\begin{figure}[t]
  \centering
  \begin{subfigure}[b]{0.48\textwidth}
    \centering
    \includegraphics[width=\linewidth]{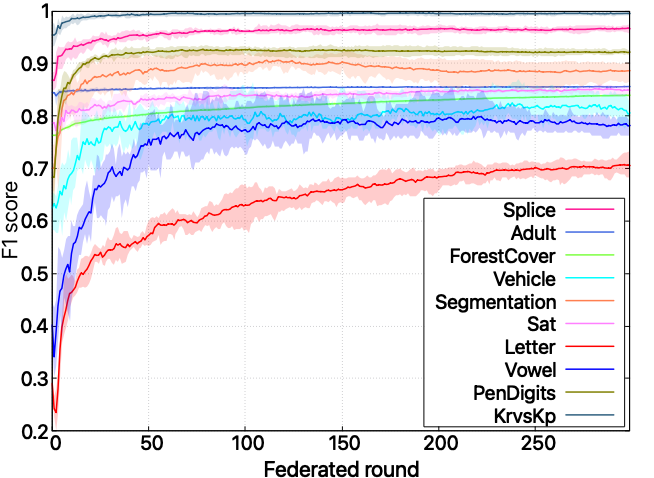}
    \caption{Aggregated AdaBoost.F model F1 score on each test set.}
    \label{fig:wandb}
  \end{subfigure}
  \hfill
  \begin{subfigure}[b]{0.48\textwidth}
    \centering
    \includegraphics[width=\linewidth]{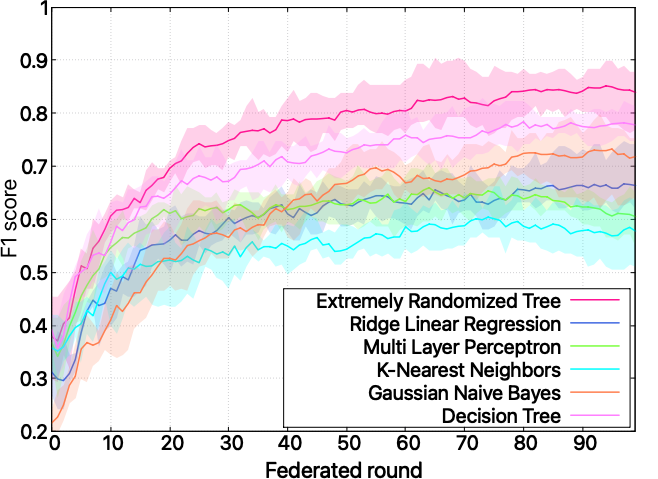}
    \caption{Example of the effect of using different ML models as weak learners.}
    \label{fig:agnostic}
  \end{subfigure}
  \caption{ML properties of \sys}
  \label{fig:ml}
\end{figure}

To demonstrate the model-agnostic property of \sys, we choose the \texttt{vowel} dataset and train different ML model types on it.
In particular, one representative ML model has been chosen from each multi-label classifier family available on SciKit-Learn: Extremely Randomized Tree (Trees), Ridge Linear Regression (Linear models), Multi-Layer Perceptron (Neural Networks), K-Nearest Neighbors (Neighbors), Gaussian Naive Bayes (Naive Bayes), and simple 10-leaves Decision Trees as baselines.
Fig.~\ref{fig:agnostic} summarises the F1 curves for the different ML models used as weak learners. 
Each model has been used out-of-the-box, without hyper-parameter tuning using the default parameters set by SciKit-Learn v1.1.2.
All ML models work straightforwardly in the proposed software without needing to code anything manually: it is sufficient to replace the class name in the experiment file.
This proves the ease with which data scientists can leverage \sys to experiment with different model types.

\subsection{Scalability Analysis}
\label{sec:scale}

We perform this scalability study using the HPC nodes from Sec.~\ref{ssec:ablation} and Monte Cimone~\cite{9908096}, the first available HPC-class cluster based on RISC-V processors. 
It comprises eight computing nodes equipped with a U740 SoC from SiFive integrating four U74 RV64GCB cores @ 1.2 GHz, 16GB RAM and a 1 Gb/s interconnection network.

We select the \texttt{forestcover} dataset for running these experiments, being the largest dataset used in this study, split into a 485K training samples and 10K testing samples.
The weak learner is the same 10-leaves SciKit-Learn Decision Tree from Sec.\ref{ssec:correctness}.
We lowered the number of federated training rounds to 100 since they are enough to provide an acceptable and stable result (10 on the RISC-V system due to the longer computational times required).
Different federations have been tested, varying numbers of Collaborators from 2 to 64 by powers of 2. 
We went no further since OpenFL is designed to suit a cross-silo FL scenario, which means a few tens of clients.
We investigated two different scenarios: \emph{strong scaling}, where we increase the collaborators while keeping the same problem size by spitting the dataset samples in uniform chunks across collaborators; and \emph{weak scaling}, where we scale the problem size with the number collaborators by assigning each collaborator the entire dataset.
In both cases, the baseline reference time is the time taken by a federation comprising the aggregator and a single collaborator.
We report the mean over 5 runs for each experiment.

\begin{figure}[t]
  \centering
  \begin{subfigure}[b]{0.45\textwidth}
    \centering
    \includegraphics[width=\textwidth]{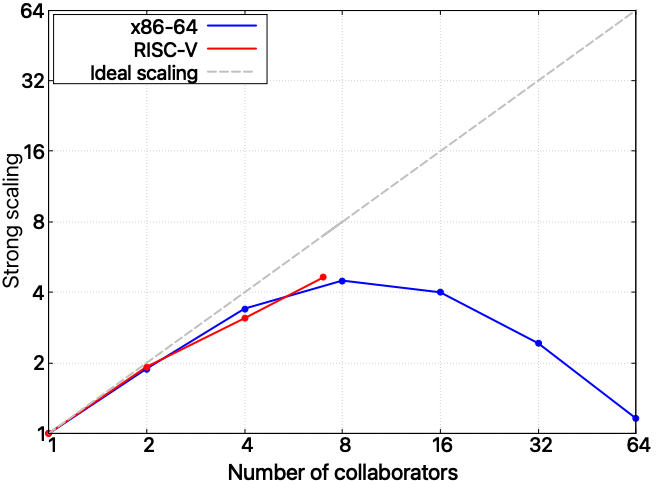}
    \caption{Strong scaling}
    \label{fig:strong}
  \end{subfigure}
  \hfill
  \begin{subfigure}[b]{0.45\textwidth}
    \centering
    \includegraphics[width=\textwidth]{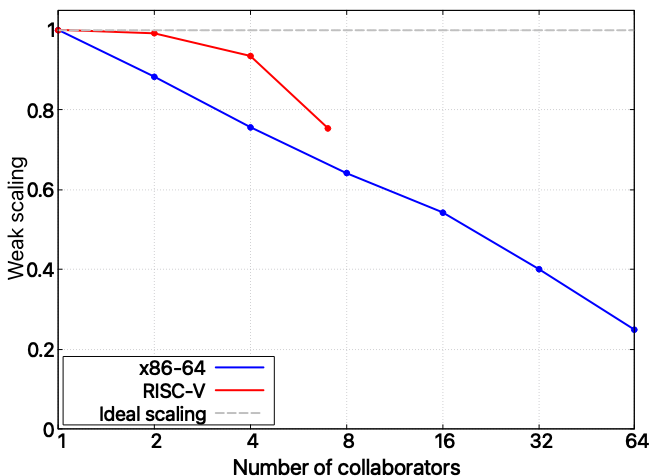}
    \caption{Weak scaling}
    \label{fig:weak}
  \end{subfigure}
  \caption{Strong and weak scaling properties of \sys.}
  \label{fig:scaling}
\end{figure}



Fig.~\ref{fig:scaling} shows the strong and weak scaling properties of \sys. 
The RISC-V plot stops at 7 because we have just 8 nodes in the cluster, and we want to avoid sharing a node between the aggregator and collaborator to maintain the same experiment system setting. 
In the strong scaling scenario, the software does not scale efficiently beyond 8 HPC nodes, as the execution becomes increasingly communication-bound. 
The same also affects the weak scaling. 
Nevertheless, the degradation is sublinear (each point on the x/axis doubles the number of nodes). 
This is important because the main benefit in the FL scenario is the additional training data brought in by each contributor node.
The RISC-V cluster exhibits better strong scalability when comparing the two clusters. 
This is justified by the slower compute speed of the RISC-V cores leading to higher training times, making the execution more compute-bound, especially for a low number of nodes. 
The weak scalability on the RISC-V cluster suffers from the lower network speed.
Since real-world cross-silo federations rarely count more than a dozen participants, it can be assessed that \sys is suitable for experimenting with such real-world scenarios.


\section{Discussion}

The implementation experience of \sys and the subsequent experimentation made it evident that current FL frameworks are not designed to be as flexible as the current research environment needs them to be.
The fact that the standard workflow of OpenFL was not customisable in any possible way without modifying the code and that the serialisation structure is DNN-specific led the authors to the idea that a new, workflow-based FL framework is needed.
Such a framework should not implement a fixed workflow but allow the user to express any number of workflow steps, entities, the relations between them, and the objects that must be exchanged.
This property implies the generalisation of the serialisation infrastructure, which cannot be limited to tensors only.
Such an approach would lead to a much more straightforward implementation of newer and experimental approaches to FL, both from the architectural and ML perspective.

Furthermore, the use of asynchronous communication can help better manage the concurrent architecture of the federation.
These systems are usually slowed down by stragglers that, since the whole system is supposed to wait for them, will slow down the entire computation.
In our experience implementing \sys, a significant part of the scalability issues is determined by the waiting time between the different collaborators taking part in the training.
While such an approach would improve the scalability performance of any FL framework, it also underlies the investigation of how to simultaneously handle newer and older updates.
This capability would improve the computational performance of gradient and non-gradient-based systems: the relative aggregation algorithms must be revised to accommodate this new logic.
This matter is not trivial and deserves research interest.
Lastly, due to the possibility of exploiting less computationally requiring models, \sys can easily be used to implement FL on low-power devices, such as systems based on the new RISC-V.
\section{Conclusions}

A model-agnostic modified version of \intel OpenFL implementing the AdaBoost.F federated boosting algorithm, named \sys, has been proposed.
The proposed software has been proven to implement the AdaBoost.F algorithm correctly and can scale sufficiently to experiment efficiently with small cross-silo federations.
\sys is open-source, freely available online, easily installable, and has a complete set of already implemented examples.
To our knowledge, \sys is the first FL framework to implement a model-agnostic, non-gradient-based algorithm. 
This effort will allow researchers to experiment with this new conception of FL more freely, pushing the concept of model-agnostic FL even further.
Furthermore, this work aims to contribute directly to the RISC-V community, enabling FL research on this innovative platform.

\subsubsection{Acknowledgments} This work has been supported by the Spoke “FutureHPC \& BigData” of the ICSC – Centro Nazionale di Ricerca in “High Performance Computing, Big Data and Quantum Computing”, funded by European Union – NextGenerationEU and the EuPilot project funded by EuroHPC JU under G.A. n. 101034126.


\bibliographystyle{splncs04}
\bibliography{Bibliography/bibliography.bib}

\end{document}